\documentclass[a4paper,fleqn]{cas-sc}
\pdfoutput=1
\usepackage[numbers,sort&compress]{natbib} 
\usepackage[linesnumbered,ruled]{algorithm2e}
\usepackage{graphicx} 
\usepackage{subfigure} 
\usepackage{hyperref}
\usepackage{siunitx}
\usepackage{arydshln} 
\usepackage{amsmath}  
\usepackage{amssymb}  
\usepackage{booktabs}
\usepackage{multirow}
\hypersetup{hidelinks} 
\usepackage{float}
\usepackage{verbatim} 
\usepackage{apalike}
\restylefloat{figure}
\restylefloat{table}
\def\tsc#1{\csdef{#1}{\textsc{\lowercase{#1}}\xspace}}
\tsc{WGM}
\tsc{QE}
\begin{document}
\let\WriteBookmarks\relax
\def\floatpagepagefraction{1}
\def\textpagefraction{.001}

% % Short title
% \shorttitle{<short title of the paper for running head>}    

% % Short author
% \shortauthors{<short author list for running head>}  

\title [mode = title]{MVMR-FS : Non-parametric feature selection algorithm based on Maximum inter-class Variation and Minimum Redundancy}  


% Title footnote 1. 题目脚注


\author[1]{Haitao Nie}[style=chinese]
\address[1]{School of Automation, Central South University, Changsha 410083, China}
\fnmark[1] %作者顺序
% \cormark[<corr mark no>]

% \fnmark[<footnote mark no>]

% \ead{<email address>}

% % URL of the first author
% \ead[url]{<URL>}

% \credit{<Credit authorship details>}

% \affiliation[<aff no>]{organization={},
% %          citysep={}, % Uncomment if no comma needed between city and postcode

\author[1]{Shengbo Zhang}[style=chinese]
% Footnote of the second author
\fnmark[2]

% % Email id of the second author
% \ead{}

% % URL of the second author

% % Credit authorship

% \affiliation[<aff no>]{organization={},
% %          citysep={}, % Uncomment if no comma needed between city and postcode

% \cortext[1]{Corresponding author}


\author[1]{Bin Xie}[style=chinese]
\cormark[1] 
\cortext[1]{Corresponding author:\ead{xiebin@csu.edu.cn}} 

\begin{abstract}
How to accurately measure the relevance and redundancy of features is an age-old challenge in the field of feature selection. However, existing filter-based feature selection methods cannot directly measure redundancy for continuous data. In addition, most methods rely on manually specifying the number of features, which may introduce errors in the absence of expert knowledge. In this paper, we propose a non-parametric feature selection algorithm based on maximum inter-class variation and minimum redundancy, abbreviated as MVMR-FS. We first introduce supervised and unsupervised kernel density estimation on the features to capture their similarities and differences in inter-class and overall distributions. Subsequently, we present the criteria for maximum inter-class variation and minimum redundancy (MVMR), wherein the inter-class probability distributions are employed to reflect feature relevance and the distances between overall probability distributions are used to quantify redundancy. Finally, we employ an AGA to search for the feature subset that minimizes the MVMR. Compared with ten state-of-the-art methods, MVMR-FS achieves the highest average accuracy and improves the accuracy by 5\% to 11\%.
\end{abstract}


\begin{highlights}
\item 
\item 
\item 
\end{highlights}

% Keywords
\begin{keywords}
 	Non-parametric feature selection \sep 
 Filter-based feature selection\sep Probability density \sep Wasserstein distance 
\end{keywords}

% maketitle原始位置
\maketitle

%%%%%%正文%%%%%%%%

\section{Introduction}\label{Indro}
The rapid development of data mining has been driven by the massive amount of high-dimensional data, which also renders many learning algorithms invalid\cite{li_feature_2018}. Therefore, feature selection(FS), which can prepare low-dimensional and understandable data, has attracted widespread attention\cite{sheikhpour_survey_2017,gheyas_feature_2010}. In the feature set, there are three types of features: relevant, irrelevant, and redundant\cite{chakraborty_selecting_2008}. Relevant features are associated with the target variable and make a significant contribution to prediction. Conversely, irrelevant features do not correlate with the target variable. Redundant features refer to highly correlated features with duplicated information. The essence of any feature selection algorithm is to identify relevant features while eliminating irrelevant and redundant ones\cite{chakraborty_feature_2015}. However, accurately measuring the relevance and redundancies within feature sets remains a challenge.\par
Filter-based supervised feature selection mostly evaluates and ranks the utility of individual feature and then selects a specified number of features\cite{li_feature_2018}. Since this process requires manually specifying the number of features, we refer to it as a parameterized feature selection framework in this paper. Methods applicable to continuous data under this framework cannot measure redundancy, and methods that can measure redundancy require discretizing the continuous data\cite{li_feature_2018}. However, discretization is prone to problems\cite{garcia_survey_2013} such as redundancy misclassification, information loss, and alteration of the decision boundaries, which are detrimental to downstream decision tasks. In addition, parametric feature selection requires domain expert knowledge to specify the number of features more accurately, otherwise it may introduce human error and fail to exploit the contribution between joint features. \par
In this paper, we select the optimal feature set from the perspective of probability distribution and propose a non-parametric feature selection algorithm based on maximum inter-class variation and minimum redundancy (MVMR-FS). Initially, we introduce both supervised and unsupervised kernel density estimation to capture the inter-class probability distribution of features and the overall probability distributions, respectively. Subsequently, we present the criteria for maximum inter-class variation and minimum redundancy (MVMR), wherein the inter-class probability distributions are employed to reflect feature relevance, and the distances between overall probability distributions are used to quantify redundancy. Finally, we employ an adaptive genetic algorithm\cite{srinivas_adaptive_1994} to search for a feature subset that minimizes the MVMR. \par
The main contributions of MVMR-FS are as follows: 
\begin{enumerate}
    \item We propose a non-parametric feature selection algorithm called MVMR-FS, which is applicable to continuous data. In contrast to previous approaches, our method eliminates the need to manually specify the number of features, thus avoiding reliance on expert knowledge and subjective human intervention. Moreover, the model without feature number restriction offers greater flexibility, enabling the exploration of optimal feature sets.  
    \item MVMR-FS can measure redundancy without discretizing continuous data. This feature allows the MVMR-FS to fully utilize the information in continuous data and more accurately measure the redundancy within the feature set, avoiding the misjudgment of redundancy caused by discretization.
    \item MRMV-FS measures the overall utility of the feature set to provide a globally optimized set rather than a combination of individual optimal feature. This metric also helps to mine interactions, complementary relationships, and joint contributions between features to avoid missing important features.
\end{enumerate}

\par
The remainder of the paper is organized as follows: Section \ref{Related Work} describes all the knowledge needed to understand this paper. Section \ref{Methods} delves into the principles and workflow of MVMR-FS. Section \ref{configu} and Section \ref{results} describe the experimental configuration and experimental results. Finally, we give a brief conclusion in the Section \ref{Conclusion}.

% 框架图注释
\begin{figure*}[ht]
    \centering
    % scale=0.17
    \includegraphics[width=\textwidth]{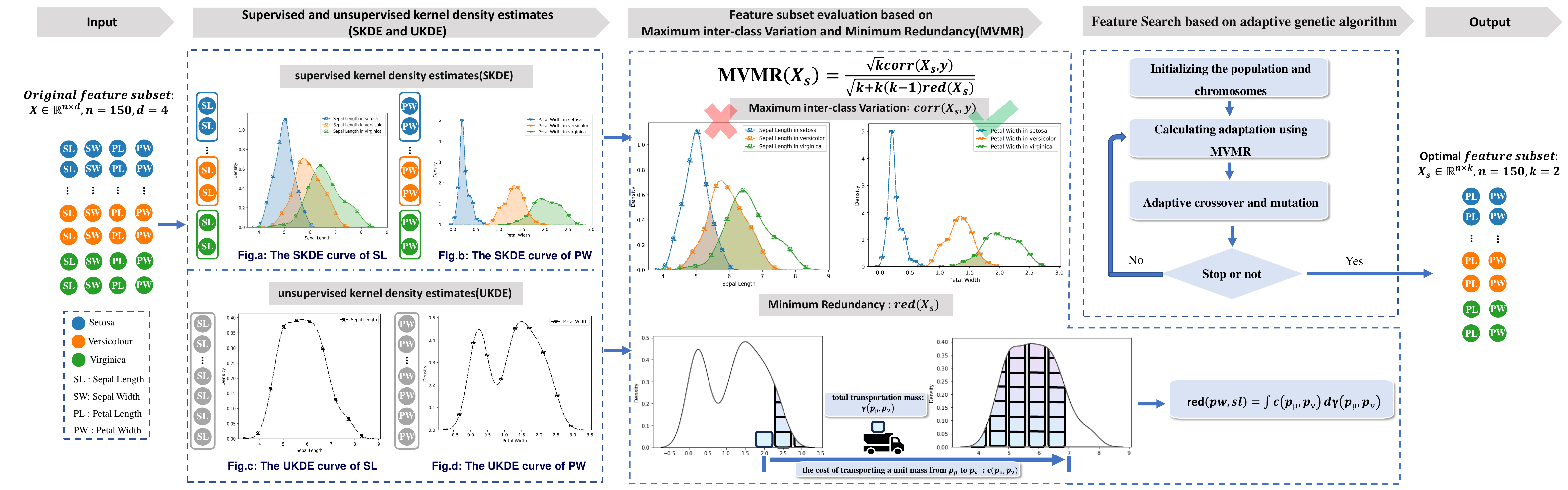}
    \caption{\textbf{An overall framework of MVMR-FS.} The input to the model is the original dataset $X \in \mathbb{R}^{n \times d}$, and the output is the best dataset $X_{s} \in \mathbb{R}^{n \times k}$ selected by MVMR-FS. First, we conduct SKDE and UKDE to capture the inter-class probability distribution of features and the overall probability distributions. Subsequently, based on the above distributions, MVMR coefficients are proposed to maximize inter-class variation while minimizing redundancy. Finally, we employ an AGA to search for the optimal feature subset $X_{s} \in \mathbb{R}^{n \times k}$.} 
    \label{framework}
\end{figure*}
\section{Related Work}
\label{Related Work}
\subsection{Filter-based feature selection methods}
Relevance and redundancy are significant concepts in feature selection. Relevance refers to the relationship between features and category information. Features that are strongly associated with the category information can be used to build precise models. Redundancy denotes the presence of highly correlated features within a feature set. The purpose of feature selection is to select a feature subset with high relevance and low redundancy.\par 
Different feature selection algorithms define various criteria to measure the relevance and redundancy within the feature set. We classify these methods into two categories: similarity-based and information-theoretic-based methods\cite{li_feature_2018}. Similarity-based methods\cite{ikonja_theoretical_2003,nie_trace_2008,sun_feature_2021} focus on the ability of features to maintain data similarity. Information-theoretic-based methods\cite{peng_feature_2005,leonardis_conditional_2006,brown_conditional_2012}, on the other hand, select features through filtering criteria designed based on information theory.\par
Similarity-based methods are easy to implement and can be applied to both continuous and discrete data\cite{sun_feature_2021}. However, Similarity-based methods cannot handle redundant features, which can lead to model overfitting. In contrast, information-theoretic-based methods can mine both relevant and redundant features, which greatly helps in building more accurate models. However, information-theoretic-based methods is only applicable to discrete data because most concepts in information theory only work with discrete data\cite{li_feature_2018}. A series of methods have been created to discretize continuous data \cite{flores_non-parametric_2022,senavirathne_rounding_2019,lin_deep_2022}. Nevertheless, discretization involves mapping continuous values to discrete values, which can result in information loss and changing the decision boundaries of the original data\cite{li_feature_2018,garcia_survey_2013}. In addition, discretization causes most of the features to have the same value, which can lead to constructing a more incomprehensible model. \par 
Reviewing the previous literature, we find that methods applicable to continuous data cannot mine redundant features. Methods that can mine redundant features cannot be applied to continuous data. Our MVMR-FS model can resolve this contradiction. MVMR-FS is a method that can be applied to continuous data as well as mining redundant features. Therefore, MVMR-FS can achieve better performance on continuous datasets.\par
\subsection{Evaluation strategies in feature selection}
The standard evaluation strategies are individual feature evaluation and feature subset evaluation. Individual feature evaluation does not consider the correlation between joint features, but relatively measures the individual utility of each feature and ranks them. Finally, the top-ranked features are manually selected as the optimal feature set. The above process can be expressed using Eq.~\eqref{individual-eq}
\begin{equation}
    ranking=(r_1,r_2,\dots,r_d)
    \label{individual-eq}
\end{equation}
$ranking$ is the ranked list obtained by applying individual feature evaluation to the feature set $X \in \mathbb{R}^{n \times d}$. $r_1$ represents the feature with the highest utility ranking, and $r_d$ represents the feature with the lowest ranking. When selecting three features, the top three rankings, $r_1$, $r_2$, and $r_3$, will be chosen. Conversely, feature subset evaluation focuses on the overall utility of the set rather than individually evaluating each feature. As shown in Eq.~\eqref{subset-eq}, $E()$ represents the utility evaluation function, and $(s_1,s_2,\dots,s_d)$ represents which features  in $X \in \mathbb{R}^{n \times d}$ are currently being evaluated.
\begin{equation}
    score=E((s_1,s_2,\dots,s_d)),s_d\in\left \{0,1  \right \} 
    \label{subset-eq}
\end{equation}
\par
Most filter-based feature selection methods\cite{ikonja_theoretical_2003,nie_trace_2008,peng_feature_2005,brown_conditional_2012,leonardis_conditional_2006} use individual feature evaluation, which can be efficient for feature selection. However, individual feature evaluation algorithms lack a comprehensive assessment of feature subsets and cannot explore the interdependencies among features, thereby often failing to select the globally optimal feature set. Moreover, these methods require artificially specifying the optimal number of features, which is highly prone to errors and degrades the accuracy of subsequent classifiers. Datasets with too many features may introduce redundant features and noise, leading to overfitting of the model. On the other hand, datasets with too few features may not provide enough information for the model to capture the underlying patterns in the data, thereby reducing the accuracy and reliability of the model.\par 
The feature subset evaluation method used in this paper has received widespread attention. In contrast to individual feature evaluation, feature subset evaluation focus on the overall utility of the subset rather than individual feature. This category of methods\cite{hall_feature_1999,yu_feature_2003} does not require human intervention and can automatically identify the optimal feature subset. Furthermore, the feature subset evaluation strategy also considers the interactions between features, aiding in exploring potential feature combinations and identifying redundant features. But the existing methods using feature subset evaluation strategies can only be applied to discrete data and not to continuous data.\par
\subsection{Wasserstein distance}
\label{related-wasserstein}
In mathematics, Wasserstein distance \cite{panaretos_statistical_2019} refers to the distance function between probability distributions on a given metric space $M$. This metric is known as the bulldozer distance in computer science. For a metric space $(M,d)$, assume that every Borel probability measure on $M$ is a Radon measure. For a finite $p \geq 1$, $P_p(M)$ denotes the set of probability measures $\mu$ for all moments of order $p$ on $M$. We can also say that $x_0$ in $M$ satisfies the following properties:

\begin{equation}
    \int_{M}d{(x,x_0)}^p\mathrm{~d}\mu(x)<\infty
    \label{x0}
\end{equation}
Therefore, the Wasserstein p-distance between the two probability measures $u$ and $v$ on $P_p (M)$ can be defined as

\begin{equation}
W_{p}(\mu, \nu):  =  \left(\inf _{\gamma \in \Gamma(\mu, \nu)} \int_{M \times M} c(x, y)^{p} \mathrm{~d} \gamma(x, y)\right)^{1 / p}
    \label{p wassers}
\end{equation}

$\Gamma(\mu, \nu)$ is the set of all measures on ${M \times M}$.We interpret the Wasserstein distance by borrowing from the optimal transport problem. We assume that the cost function for transporting distribution $\mu(x)$ to $\nu(x)$ is $c(x,y)$, and the mass of the transport is $\gamma(x, y)$. The mass of the transport $\gamma(x, y)$ must satisfy:

\begin{equation}
    \int \gamma(x,y)\mathrm{~d}y=\mu(x)
    \label{tiaojian1}
\end{equation}

\begin{equation}
    \int \gamma(x,y)\mathrm{~d}x=\nu(x)
    \label{tiaojian2}
\end{equation}
Eq.~\eqref{tiaojian1} indicates that the total mass moving from point $x$ to the other points must equal the total mass at point $x$ before the move. Eq.~\eqref{tiaojian2} indicates that the final total mass at point $y$ equals the total mass moving from the other points to point $y$. Thus the total cost of transportation is
\begin{equation}
    \iint c(x,y)\gamma (x,y)\mathrm{~d}(x)\mathrm{d}(y)=\int c(x,y)\mathrm{~d}\gamma (x,y)   
    \label{zongdaijia}
\end{equation}

The transportation solution is not unique, and the optimal transportation solution is the one with the lowest total cost among all possible solutions. The cost of the optimal transportation solution is:
\begin{equation}
C=\inf _{\gamma \in \Gamma(\mu, \nu)} \int c(x, y) \mathrm{~d} \gamma(x, y)
    \label{1-wassers}
\end{equation}
If the cost of a move is simply the distance between the two points, then the optimal cost is identical to the definition of the wasserstein 1-distance.
\section{Methodology}
\label{Methods}
Feature selection employs measurement criteria and search algorithms to select more beneficial features for downstream tasks. However, how to perform both relevance and redundancy measures on continuous data and construct non-parametric feature selection algorithms remain considerable challenges in this field.\par 
The MVMR-FS proposed in this paper can solve the above problems. We first perform supervised and unsupervised kernel density estimation on the features to capture their similarities and differences in inter-class and overall distributions. The above information is then utilized to construct MVMR coefficients that include variation and redundancy terms. The variation term evaluates the contributions of features based on inter-class probability distributions, while the redundancy term uses transportation theory to measure the similarity between features. Finally, we employ an AGA to search for the feature subset that minimizes the MVMR. The workflow of MVMR-FS is illustrated in Fig.~\ref{framework}.
\par
\subsection{Probability density functions of the features}
\label{methods-kde}
The probability density function provides information on the central tendency, dispersion, kurtosis and tail distribution of the data distribution, which helps us to determine feature importance and redundancy. \par 
However, we generally do not know the probability density functions of the features, so we employ Kernel Density Estimation (KDE) for probability density estimation. Let $\{x_1,x_2, \dots, x_m\}$ be the $m$ independent random variables drawn from an unknown density distribution $f_X (x)$. The distribution of $m$ independent random variables is obtained using KDE as follows:
\begin{equation} 
  \hat{f}_h(x)=\frac{1}{mh}\sum_{i=1}^{m}K(\frac{{x-X_i}}{h})
  \label{KDE}
\end{equation}
$h$ is the bandwidth, and $K()$ is a non-negative kernel function. $X_i$ denotes the $i\text{-}th$ variable.\par
The real-world dataset is likely to have outliers and noise,  so we choose a more robust Gaussian kernel. In addition, the Gaussian kernel is also more capable of capturing the overall trend and distribution characteristics of the data. The formula for the Gaussian kernel is as follows:
\begin{equation} 
  K(\nu )=\frac{1}{\sqrt{2 \pi}} e^{\frac{-\nu ^{2}}{2}}
  \label{GB kernel}
\end{equation}
\par
When performing feature selection, we have an original feature set $X \in \mathbb{R}^{n \times d}$ with $n$ samples and $d$ features. The samples in $X \in \mathbb{R}^{n \times d}$ belong to $c$ classes. We perform supervised kernel density estimation and unsupervised kernel density estimation on $X \in \mathbb{R}^{n \times d}$.\par
\subsubsection{Supervised and unsupervised kernel density estimates}
 \label{SKDE}
The key distinction between supervised and unsupervised kernel density estimation lies in the utilization of category information. Supervised Kernel Density Estimation (SKDE) initially divides the feature set $X \in \mathbb{R}^{n \times d}$  into $c$ subsets based on category information, where each subset contains samples belonging to the same class. We denote the $c$ features subsets as: $M=\{C_1,C_2,\ldots,C_c,(C_i \subseteq \mathbb{R}^{v_i \times d})\}$. Subsequently, kernel density estimation is independently performed on each subset. The supervised density functions of the $j\text{-}th$ feature in the subset $C_i$ is denoted as $s_j^i(x)$. In contrast, unsupervised kernel density estimation (UKDE) directly estimates the density function of the features in the $X \in \mathbb{R}^{n \times d}$ without using category information. The unsupervised density function of the $j\text{-}th$ feature is denoted as $u_j(x)$. \par
SKDE and UKDE provide insights into the density distribution, central tendency, skewness, and kurtosis of data from different perspectives. SKDE focuses on the similarities and differences between categories, providing information on whether features contribute to classification. On the other hand, UKDE emphasizes the overall distribution of features and helps measure the interdependencies among them.\par
We illustrated the concepts mentioned above by plotting Fig.~\ref{framework} using the information from the Iris dataset\footnote{\url{https://archive.ics.uci.edu/dataset/53/iris}}. The Iris dataset comprises measurements of 150 iris flowers belonging to three species: setosa, versicolor, and virginica. Setosa, versicolor, and virginica are represented by blue, orange, and green circles, respectively. Each sample has measurements of four features, including sepal length(SL), sepal width(SW), petal length(PL), and petal width(PW).\par 
Subfigures a and b in Fig.~\ref{framework} depict the SKDE curves. Subfigures c and d represent the UKDE curves. Comparing subfigures a and b, we observe that the SL exhibits less variability across categories than the PW. By observing Subfigure b, we can identify that the SKDE curve for the setosa exhibits the steepest slope. Furthermore, we observe that the PW has a higher skewness in the Versicolour compared to the Virginica. The differences in features across categories directly reflect the ability to classify the target accurately. The UKDE curves provide information on the overall distribution of features. Subfigure c demonstrates that the SL follows a unimodal distribution, while subfigure d indicates that the PW exhibits a bimodal distribution. The differences in distribution shapes also indicate that the SL and PW features are dissimilar, suggesting they are not redundant.\par
\subsection{Feature subset evaluation based on Maximum inter-class Variation and Minimum Redundancy 
}
\label{methods-PDE}
Most feature selection algorithms use the individual feature evaluation strategy, but this strategy has two issues: 1. There is no guarantee that the performance of the feature subset consisting of the optimal single features is also excellent. 2. It is difficult for humans to determine the optimal number of features, and selecting too many or too few features can lead to issues such as decreased model accuracy, overfitting, and underfitting.\par 
Our proposed feature subset evaluation criterion based on Maximum inter-class Variation and Minimum Redundancy (MVMR) can solve the above issues. MVMR is formulated as:

\begin{equation}
     MVMR(X_s)=\frac{\sqrt{k}corr(X_s,y)}{\sqrt{k+k(k-1)red(X_s)}}
     \label{PDE-equation}
 \end{equation}
 MVMR-FS utilizes Eq.~\eqref{PDE-equation} to quantify the relevance and redundancy within the feature set $X_s \in \mathbb{R}^{n \times k}$. $k$ represents the number of selected features. $corr(X_s,y)$ denotes the relevance between $X_s$ and the category information. $red(X_s)$ denotes the redundancy within $X_s \in \mathbb{R}^{n \times k}$. The result of $MVMR(X_e)$ is a non-negative value. A smaller $MVMR(X_e)$ value indicates that the current feature subset $X_s \in \mathbb{R}^{n \times k}$ is more favorable for the classification task. $corr(X_s,y)$ and $red(X_s)$ are shown in Eq.~\ref{corr(X,y)} and Eq.~\ref{corr(x_i,x_j)}.
 
 \begin{equation}
     corr(X_s,y)=\frac{\sum_{i=1}^{k}sim(x_i,y)}{k}
     \label{corr(X,y)}
 \end{equation}

 \begin{equation}
     red(X_s)=\frac{\sum_{i=1}^{k-1} \sum_{j=i+1}^{k}red(x_i,x_j)}{k}
     \label{corr(x_i,x_j)}
 \end{equation}
 $x_i$,$x_j$ are the features in the feature set $X_s \in \mathbb{R}^{n \times k}$. $sim(x_i,y)$ denotes the relevance measure for $x_i$. $red(x_i,x_j)$ denotes the redundancy measure for $x_i$ and $x_j$. We will introduce $sim(x_{i},y)$ and $red(x_{i},x_{j})$ in Section.~\ref{methods-Similarity metric} and Section~\ref{methods-Redunacy}, respectively.\par
 \subsubsection{Relevance metrics based on Maximum inter-class Variation}
 \label{methods-Similarity metric}
 The relevance metric is used to identify features that help with classification. Features can provide valuable classification information when they exhibit significant differences across categories. MVMR-FS utilize the inter-class probability distribution of features, also known as the supervised density functions $s_j^i(x)$ in Section.\ref{SKDE}, to construct $sim(x_i,y)$ for evaluating the variations. The formula for $sim(x_i,y)$ is as follows: \par 
 \begin{equation} 
sim(x_{i},y)=\frac{\int_{c}^{d}f_{overlap}(x)\,\mathrm{d}x}{\int_{a}^{b}f_{outer}(x)\,\mathrm{d}x} 
  \label{sim-next-max}
\end{equation}
\par
The formulas for $f_{outer}(x)$ and $f_{overlap}(x)$ in Eq.~\eqref{sim-next-max} are as follows:
\begin{equation} 
    f_{outer}(x)=\max{(s_{j}^{1}(x),s_{j}^{2}(x),\dots,s_{j}^{c}(x))},x\in[a,b]
    \label{outer_edge}
\end{equation}

\begin{equation} 
    f_{overlap}(x)=\text{second} {(s_{j}^{1}(x),s_{j}^{2}(x),\dots,s_{j}^{c}(x))},x\in[c,d]
    \label{overlap_edge}
\end{equation}
$\max(x)$ and $\text{second}(x)$ denote taking the maximum and second largest value, respectively. $sim(x_i,y)$ represents the ratio of the intersection area to the union area under the probability density curves of the categories, which can be interpreted as measuring the degree of overlap in the probability density curves. Therefore, $sim(x_i,y)$ provides information about the discriminative power of features and the separability of categories. If the probability density curves of features exhibit low overlap, resulting in a lower value of $sim(x_i,y)$, it indicates that the current feature can effectively separate the categories. The value range of $sim(x_i, y)$ is [0, 1]. Features with lower $sim(x_i,y)$ values can provide more information for classification.\par 
We illustrate the above concept using Fig.a and Fig.b in Fig~\ref{framework}. The numerator of $sim(x_i, y)$ is the total area of the overlapping colored parts, and the denominator is the total area of the colored parts. Compared to PW, the SKDE curves of SL are more similar and have a larger overlap ratio. So we believe using feature SL will build a classifier with lower performance. The results in Experiment.~\ref{The nature of PDE} and Table.~\ref{exper1} confirm our idea. The accuracy of the classifier constructed using the PW feature is 1.0000, while the accuracy of the classifier constructed using the SL feature is only 0.7193.
\subsubsection{Redundancy metrics based on Minimum Redundancy}
\label{methods-Redunacy}
Redundancy metrics are an essential step in feature selection. Some methods may not be able to detect redundant features, which can lead to model overfitting and the curse of dimensionality\cite{ikonja_theoretical_2003}. Redundancy refers to the intercorrelation between features, which is quantified by MVMR-FS using the Wasserstein distance. The Wasserstein distance computes the minimum transformation distance between probability distributions. Therefore, the more similar the features, the smaller the Wasserstein distance between them. The formula for measuring redundancy is as follows:

\begin{equation}
red(x_i,x_j)=\inf _{\gamma \in \Gamma(\mu, \nu)} \int c(p_{\mu}, p_{\nu})\, \mathrm{d} \gamma(p_{\mu}, p_{\nu})
    \label{red_m}
\end{equation}

 $\mu$, $\nu$ are the probability density functions of the features $x_i$, $x_j$. In our paper, $\mu$ and $\nu$ are obtained by the unsupervised kernel density estimation described in Section.\ref{SKDE}. $p_{\mu}$ and $p_{\nu}$ denote points on $\mu$ and $\nu$, respectively. The cost function for transporting distribution $\mu(x)$ to $\nu(x)$ is $c(p_{\mu}, p_{\nu})$, and the mass of the transport is $\gamma(p_{\mu}, p_{\nu})$. The detailed explanation of Eq.~\eqref{red_m} is described in the Section.~\ref{related-wasserstein}.\par
To measure redundancy within the set accurately, we need to normalize the features before calculating Eq.~\eqref{red_m}. If normalization is not performed, the redundancy measure will not accurately assess the similarity between some features. For example, features with proportional relationships may differ significantly in their probability density curves, but the information provided by these features is highly similar. This similarity can be precisely mined by the redundancy measure only after normalization. We also verify the necessity of normalization in Experiment.~\ref{The nature of PDE}.\par
In comparison to other distance metrics such as euclidean distance and Kullback-Leibler divergence\cite{mollersen_data-independent_2016}, Wasserstein distance has several advantages, including comprehensive consideration of distribution shape, insensitivity to scaling transformations, and desirable mathematical properties\cite{panaretos_statistical_2019}. Traditional distance metrics\cite{cha2007comprehensive} like euclidean distance only consider the distance between the centers or means of distributions, neglecting important information about the distributions shape, patterns, and arrangement. On the other hand, Wasserstein distance provides a more accurate metric for distributions with significant differences or multimodality by considering shape and transportation costs. This also aids MVMR-FS in more effectively measuring the redundancy among features.
\subsection{Feature Search based on adaptive genetic algorithm}
\label{methods-AGA}
The choice of search algorithm is critical to the performance and effectiveness of feature selection. Common search strategies include exhaustive search, heuristic search, and random search. 
Exhaustive search is rarely used due to its high computational complexity. Heuristic search\cite{wang_feature_2016} is simple and fast to implement but is prone to getting trapped in local optima. 
Therefore, we use the adaptive genetic algorithm\cite{srinivas_adaptive_1994} belonging to random search.\par 
When performing feature selection on the set $X \in \mathbb{R}^{n \times d}$ with $n$ samples and $d$ features, we initialize a population containing $N$ binary chromosomes. The length of each chromosome is $d$. The numbers on the chromosome represent whether the corresponding feature is selected or not. MVMR(Eq.~\eqref{PDE-equation}) was used to calculate the fitness values of the chromosomes. We employed binary tournaments to select individuals for adaptive crossover and mutation. The adaptive crossover probability $p_c$ and mutation probabilities $p_m$ are as follows  :
% 自适应交叉概率
\begin{equation}
    \resizebox{0.90\hsize}{!}{$
        p_c=
    \begin{cases}
    \mathrm{p_{c-max}}-\frac{\mathrm{(p_{c-max}-p_{c-min})}(f_{avg}-f_{c})}{f_{avg}-f_{min}} &, f_{c}\le f_{avg}\\
    \mathrm{p_{c-max}} &, f_{c} > f_{avg}
    \end{cases}
    $}
    \label{jiaocha}
\end{equation}

% 自适应变异概率
\begin{equation}
    \resizebox{0.90\hsize}{!}{$p_{m}=
    \begin{cases}
    \mathrm{p_{m-max}}-\frac{\mathrm{(p_{m-max}-p_{m-min})}(f_{avg}-f_{m})}{f_{avg}-f_{min}} &, f_{m}\le f_{avg}\\
    \mathrm{p_{m-max}} &, f_{m} > f_{avg}
    \end{cases}$}
    \label{bianyi}
\end{equation}
The minimum and average fitness values of the populations are denoted by $f_{min}$ and $f_{avg}$, respectively. $f_{c}$ denotes the smaller of the two individual fitness values to be crossed, and $f_{m}$ represents the fitness value of the individual to be mutated. $\mathrm{p_{c-max}}$ and $\mathrm{p_{c-min}}$ represent the maximum and minimum crossover probabilities, respectively. $\mathrm{p_{m-max}}$ and $\mathrm{p_{m-min}}$ represent the maximum and minimum mutation probabilities, respectively. We subsequently recalculate the fitness of the population to update the best individual. When the best individual remains unchanged for $M$ consecutive generations, the search algorithm stops and outputs the optimal feature set.\par
we consider individuals with low fitness values to be superior. For individuals with high fitness values, we should use larger crossover and mutation probabilities to increase the population's diversity. For individuals with low fitness values, we should use smaller crossover and variation probabilities to prevent destroying the optimal solution and speed up the convergence rate. The adaptive genetic algorithm's dynamic search strategy helps MVMR-FS find a better feature subset. \par 
\section{Experimental configuration}
\label{configu}
\subsection{Datasets}
% cankaowenxian
Liu \cite{li_feature_2018} published the datasets\footnote{\url{https://jundongl.github.io/scikit-feature/datasets.html}} commonly used in the feature selection. We choose eight of the open-source datasets for our experiments. These eight datasets involve Face Image Data, Biological Data, Artificial Data, and Hand Written Image Data.\par
We also modify the Iris dataset to validate the nature of MVMR. We refer to the modified Iris dataset as Artificial Iris. The Artificial Iris has five features, namely, sepal length(SL), sepal width (SW), petal length (PL), petal width (PW) and twice sepal-length(2SL). 2SL is the feature that enlarging SL by a factor of two. Information on all datasets is presented in Table \ref{Datasets}.
%数据集介绍
\begin{table}[]
\centering %居中
    \caption{Datasets used in the experiments}
    \label{Datasets}
    \resizebox{\textwidth}{!}{
\begin{tabular}{cccccc}
    % \caption{Datasets used in the experiments}
   \toprule %顶部线
   & Dataset & Instances & Features & Classes \\
   \midrule %中部线  \footnotemark
   1 & USPS   & 9285 & 256 & 10  \\
   2 & GLIOMA  & 50 & 4434 & 4  \\
   3 & lung  & 203 & 3312 & 5  \\
   4 & madelon  & 2600 & 500 & 2  \\
   5 &COIL20  & 1440 & 1024 & 20 \\
   6 & TOX\_171  & 171 & 5748 & 4 \\
   7 & warpPIE10P  & 210 & 2420 & 10\\
   8 & Prostate\_GE  & 102 & 5966 &2 \\
   9 & Artificial Iris &  150 & 5    &3\\
   \bottomrule %底部线
\end{tabular}
}
\end{table}

\subsection{Baselines}
\label{experimental-confi-baselines}
To verify the validity of MVMR-FS, we compared MVMR-FS with ten methods. Information on the ten methods is shown in Table \ref{baselines}. The description of Table \ref{baselines} is as follows:
\begin{enumerate}
    \item Metrics: The ticks in the Relevance and Redundancy columns indicate whether the corresponding method can measure relevance or redundancy.
    \item Data: A tick in the continuous column indicates that the corresponding method can be applied directly to continuous data.
    \item Evaluation strategies: The ticks in the Ranking and Subset columns indicate whether the corresponding method uses individual feature evaluation or feature subset evaluation.
\end{enumerate}

%对比方法
\begin{table}[]
\centering %居中
    \caption{The nature of the baselines}
    \label{baselines}
    \resizebox{\textwidth}{!}{
\begin{tabular}{ccccccc} 
\toprule 
&\multirow{2}{*}{Methods} 
& \multicolumn{2}{c}{Metrics} & Data&\multicolumn{2}{c}{Evaluation strategies}\\ 
\cmidrule{3-7}
& & Relevance & Redundancy&Continuous &Ranking &Subset \\ 
\midrule %中部线
1&Fisher Score\cite{sun_feature_2021}   & $\checkmark$ &   & $\checkmark$ &$\checkmark$ &\\ 
2&ReliefF\cite{ikonja_theoretical_2003} & $\checkmark$ &   &$\checkmark$ &$\checkmark$& \\

3&Trace Ratio\cite{nie_trace_2008} & $\checkmark$ &   &$\checkmark$ &$\checkmark$& \\

4&mRMR\cite{peng_feature_2005} & $\checkmark$ & $\checkmark$  & &$\checkmark$& \\
5&CIFE\cite{leonardis_conditional_2006} & $\checkmark$ & $\checkmark$  & &$\checkmark$& \\
6&JMI\cite{brown_conditional_2012} & $\checkmark$ & $\checkmark$  & &$\checkmark$& \\
7&CMIM\cite{wang_feature_2004} & $\checkmark$ & $\checkmark$  & &$\checkmark$& \\
8&DISR\cite{10.1007/11732242_9} & $\checkmark$ & $\checkmark$  & &$\checkmark$& \\
9&FCBF\cite{yu_feature_2003} & $\checkmark$ & $\checkmark$  & & &$\checkmark$ \\
10&CFS\cite{hall_feature_1999} & $\checkmark$ & $\checkmark$  & & & $\checkmark$\\
\bottomrule 
\end{tabular}
}
\end{table}

\subsection{Evaluation metrics}
We use Accuracy (acc), Variance ($\sigma^{2}$) and Pearson correlation coefficient for the experimental evaluation.\par
acc is defined as follows:
\begin{equation}
    acc=\frac{true}{all}
    \label{acc}
\end{equation}
$true$ indicates the number of correct classifications. $all$ indicates the total number of instances involved in the classification task. Variance is defined as follows:

\begin{equation}
    \sigma^{2}=\frac{\sum_i^n{(x_i-\overline{x})}^2}{n}
    \label{Var}
\end{equation}
$n$ is the number of samples. $\overline{x}$ denotes the mean of all samples. We use acc to evaluate the performance of feature selection algorithms and $\sigma^2$ to assess their stability. \par
Pearson correlation coefficient is defined as follows
\begin{equation}
    \rho_{X,Y}=\frac{cov(X,Y)}{\sigma_X \sigma_ Y}
    \label{pierxun}
\end{equation}
$cov(X,Y)$ denotes the covariance of $X$ and $Y$. $\sigma_X$ and $\sigma_ Y$ denote standard deviation. The Pearson correlation coefficient measures the magnitude of the correlation between two variables. The larger the absolute value of the Pearson correlation coefficient, the higher the correlation between the two variables.
\subsection{Operational details}
\label{Operational details}
In all experiments, the training and test sets are divided in the ratio of 8:2. We train the feature selection model on the training set and then select the features on the test set. Finally, we perform the classification tasks using K-Nearest Neighbors (KNN)\cite{altman_introduction_1992}, Gaussian Naive Bayes (GB)\cite{domingos_optimality_1997}, and Decision Trees (DT)\cite{quinlan_induction_1986}. \par We report the average accuracy and variance of the three classifiers in experiments. The high accuracy indicates that the feature selection algorithm can select excellent features. The low variance indicates that the feature selection algorithm picks features well-suited to various classifiers.  All experimental results are rounded to four decimal places.\par
The parameter settings about KNN, KDE and AGA are shown in Table \ref{parm}. Models 1-8 in Table.~\ref{baselines} are set to select 30 features on the USPS dataset and 50 features on the remaining datasets(Datasets 2-8 in Table.~\ref{Datasets}). For a fair comparison, we set methods 9, 10, and our model to select at most 30 features on Lung dataset and at most 50 features on the remaining dataset(Datasets 2-8 in Table.~\ref{Datasets}).

%参数设置
\begin{table}[]
\centering %居中
    \caption{\textbf{Parameter settings in the experiments.} $K$ denotes the number of nearest neighbors in KNN. $bandwidth$ is a parameter in KDE(Section.~\ref{methods-kde}). $M$ is the unevolved algebra in the MVMR-FS(Section.~\ref{methods-AGA}). $p_{c-max}$, $p_{c-min}$, $p_{m-max}$, $p_{m-min}$ are the adaptive crossover and mutation probabilities(Section.~\ref{methods-AGA}) in AGA.}
    \label{parm}  
    \resizebox{\textwidth}{!}{
\begin{tabular}{ccccccc}
    % \caption{Datasets used in the experiments}
   \toprule %顶部线
   $k$  &$bandwidth$& $M$&$p_{c-max}$ & $p_{c-min}$& $p_{m-max}$ & $p_{m-min}$\\
   \midrule %中部线
   3  &1 & 200&0.90 & 0.50 & 0.10 &0.01\\
   \bottomrule %底部线
\end{tabular}
}
\end{table}

\section{Experimental results}
\label{results}

\subsection{The nature of MVMR}
\label{The nature of PDE}
To verify the conjecture in Section \ref{methods-Similarity metric} and \ref{methods-Redunacy}, we reveal the effectiveness of MVMR and the necessity of normalization on the Artificial Iris dataset. The validity of MVMR is demonstrated by the correlation between the results of MVMR after normalizing the feature and classification accuracy. The necessity of normalization is argued by comparing the correlation between MVMR and classification accuracy before and after normalizing the feature.\par
Table \ref{exper1} shows the accuracies of classification using pairwise features. The results at the diagonal correspond to classifiers that simultaneously use two identical features. Table~\ref{guiyihuadebiyaoxing} and Table~\ref{experimental-guiyihuaqian} represent the metrics of MVMR on the normalized features and the unnormalized features, respectively. When measuring the features before normalization, the Pearson correlation coefficient between MVMR(Table.~\ref{experimental-guiyihuaqian}) and classification accuracy(Table.~\ref{exper1}) is -0.7422 (P-value=\num{2.1609e-05}). However, for the normalized features, the Pearson coefficient between MVMR(Table.~\ref{guiyihuadebiyaoxing}) and classification accuracy reached -0.9065 (P-value=\num{4.3952e-10}), showing an increase of 0.1643 compared to the previous value. The higher correlation coefficient and lower P-value indicate that MVMR provides more precise measurements on the normalized features.
% 实验一的分类精度
\begin{table}[]
\centering
\caption{\textbf{Classification accuracies using pairwise features.}  For example, the number 0.7632 in the second column of the first row is the accuracy of the classifier constructed using SL and SW features.}
\label{exper1}
\resizebox{\textwidth}{!}{
\begin{tabular}{cccccc}
\toprule
    & SL     & SW     & PL     & PW     & 2SL    \\ 
    \midrule
SL  & 0.7193 & 0.7632 & 0.9561 & 0.9474 & 0.7193 \\
SW  & 0.7632 & 0.5351 & 0.9825 & 1.0000      & 0.7544 \\
PL  & 0.9561 & 0.9825 & 0.9737 & 1.0000      & 0.9561 \\
PW  & 0.9474 & 1.0000      & 1.0000      & 1.0000      & 0.9474 \\
2SL & 0.7193 & 0.7544 & 0.9561 & 0.9474 & 0.7193 \\ \bottomrule
\end{tabular}}
\end{table}

%PDE结果归一化后
\begin{table}[]
\centering
\caption{\textbf{The results of MVMR for normalized pairwise features.} The correlation between the results of MVMR for normalized pairwise features(Table.~\ref{guiyihuadebiyaoxing}) and classification accuracies(Table.~\ref{exper1}) is -0.9065 (P-value=\num{4.3952e-10}).}
\label{guiyihuadebiyaoxing}
\resizebox{\textwidth}{!}{
\begin{tabular}{cccccc}
\toprule
    & SL             & SW             & PL             & PW             & 2SL            \\\midrule
SL  & 0.1215 & 0.1649 & 0.0442  & 0.0417 & 0.1215 \\
SW  & 0.1649 & 0.2309 & 0.0725 & 0.0694 & 0.1649 \\
PL  & 0.0442 & 0.0725 & 0.0072 & 0.0060  & 0.0442 \\
PW  & 0.0417 & 0.0694 & 0.0060  & 0.0052 & 0.0417 \\
2SL & 0.1215 & 0.1649 & 0.0442 & 0.0417 & 0.1215 \\
\bottomrule
\end{tabular}}
\end{table}

%MVMR结果归一化前
\begin{table}[]
\centering
\caption{\textbf{The results of MVMR for unnormalized pairwise features.} The correlation between the results of MVMR for unnormalized pairwise features(Table.~\ref{experimental-guiyihuaqian}) and classification accuracies(Table.~\ref{exper1}) is -0.7422 (P-value=\num{2.1609e-05}).The absolute value of the correlation between MVMR and classification accuracies before normalization(Table.~\ref{experimental-guiyihuaqian} and Table.~\ref{exper1}) is 0.1643 lower than that after normalization(Table.~\ref{guiyihuadebiyaoxing} and Table.~\ref{exper1}).}
\label{experimental-guiyihuaqian}
\resizebox{\textwidth}{!}{
\begin{tabular}{cccccc}
\toprule
    & SL             & SW             & PL             & PW             & 2SL            \\\midrule
SL  & 0.1215 & 0.0577 & 0.0190 & 0.0103 & 0.0237 \\
SW  & 0.0577 & 0.2309 & 0.0394 & 0.0331 & 0.0242 \\
PL  & 0.0190  & 0.0394 & 0.0072 & 0.0022  & 0.0071 \\
PW  & 0.0103 & 0.0331 & 0.0022  & 0.0052 & 0.0053 \\
2SL & 0.0237 & 0.0242 & 0.0071 & 0.0053 & 0.1215 \\
\bottomrule
\end{tabular}}
\end{table}

\subsection{Ablation study}
In Table \ref{xiaorong}, we show ablation results to verify the contribution of each component in our model. The acc, $\sigma^{2}$ and the dimensions of the feature subset are shown in Table \ref{xiaorong}. \par
To investigate the effectiveness of MVMR for feature evaluation, we compare the performance of Model 1 and Model 2. Model 2 utilizes the MVMR criterion, whereas Model 1 only uses the relevance measure from MVMR. Model 2 achieved higher classification accuracies on all datasets. The largest difference in classification accuracies is found on the GLIOMA, with model 2 improving 0.2333 over model 1. Model 2 not only improves the accuracy but also reduces the dimensionality of the feature subsets. For example, on the Prostate\_GE dataset, Model 2 reduces the feature set from 50 to 13 dimensions. The results indicate that using MVMR for feature metrics is a more effective way compared with only performing relevance metrics. \par
To explore the superiority of feature search using the adaptive genetic algorithm, we compare the performance of Model 2 and MVMR-FS. The search strategy of Model 2 is sequential forward search, while MVMR-FS uses an adaptive genetic algorithm. MVMR-FS achieved higher classification accuracy on all datasets. The largest difference in classification accuracies is found on the Prostate\_GE, and MVMR-FS improved 0.1099 compared with Model 2. MVMR-FS not only reduced the dimensionality of the feature subset but also improved the accuracy on the GLIOMA and madelon datasets. Although MVMR-FS increased the feature subset from 13 to 35 dimensions in the Prostate\_GE dataset, it also improved the accuracy by 0.1099. The results show that the adaptive genetic algorithm is a more practical feature search method.\par

%消融实验结果 之前有*
\begin{table*}[]
\centering
\caption{\textbf{Ablation study.} We perform ablation experiments in the GLIOMA, madelon, and Prostate\_GE datasets. 'MVMR' refers to our proposed Maximum inter-class Variation and Minimum Redundancy criterion(Eq.~\eqref{corr(x_i,x_j)}). 'sim' refers to the relevance measure term(Eq.~\eqref{corr(X,y)}) in the MVMR. 'SFS' denotes the sequential forward search. 'AGA' denotes the adaptive genetic
algorithm in Section.~\ref{methods-AGA}}
\label{xiaorong}
% 0.96\columnwidth
\resizebox{\textwidth}{!}{
    \begin{tabular}{c|cccc|ccc|ccc|ccc}
\hline
\multirow{2}*{Model} & \multicolumn{4}{c|}{Model Description}&\multicolumn{3}{c|}{GLIOMA} & \multicolumn{3}{c|}{madelon} &\multicolumn{3}{c}{Prostate\_GE}\\ 
& sim & MVMR & SFS & AGA & acc & $\sigma^{2}$ & dimensions & acc & $\sigma^{2}$ & dimensions &acc &$\sigma^{2}$ &dimensions\\ \hline
1 & $\checkmark$ &  & $\checkmark$ & & 0.5667 & 0.0089 & 50 & 0.7269 & 0.0051 & 50 & 0.7937 & 0.0035 & 50 \\ 

2 & $ $ & $\checkmark$ &$\checkmark$ & & 0.8000 & 0.0200  & 43 & 0.7558 & 0.0089 & 22  & 0.8413 & 0.0156 & 13 \\  
% \hline
MVMR-FS & $ $ & $\checkmark$ & &$\checkmark$ & 0.8467 & 0.0167  & 41 & 0.7788 & 0.0099 & 15 &0.9512 & 0.0005 & 35 \\
\hline
\end{tabular}
}
\end{table*}

\subsection{Comparison with State-of-the-Art Methods}
\label{results-SOTA}

\begin{table*}[]
\centering
\caption{\textbf{State-of-the-art comparison for individual feature evaluation strategy.} Methods utilizing individual feature evaluation strategies focus on the utility of individual feature. The 'Dimension Settings' represents the manually specified number of features. The 'avg\underline{ }acc' represents the average accuracy of the three classifiers(KNN,GB,DT,Section.~\ref{Operational details}). The '$\sigma^{2}$' represents the variance of the three classification accuracies. The 'mean' row represents the average of various metrics. The best accuracy in each row is marked with bold. Our model (MVMR-FS) obtains the best performance in six of the eight datasets while also obtaining the highest average accuracy.}
\label{SOTA-ranking}
\resizebox{\textwidth}{!}{
    \begin{tabular}{c|cc|cc|cc|cc|cc|cc|cc|cc|c|ccc}
\hline
\multirow{2}*{Dataset} &\multicolumn{2}{c|}{Fisher Score} & \multicolumn{2}{c|}{ReliefF} &\multicolumn{2}{c|}{Trace Ratio}&\multicolumn{2}{c|}{mRMR}&\multicolumn{2}{c|}{CIFE}&\multicolumn{2}{c|}{JMI}&\multicolumn{2}{c|}{CMIM}&\multicolumn{2}{c|}{DISR} & Dimension &\multicolumn{3}{c}{ours model(MVMR-FS)}\\ 
 & avg\underline{ }acc & $\sigma^{2}$ & avg\underline{ }acc & $\sigma^{2}$  & avg\underline{ }acc & $\sigma^{2}$ & avg\underline{ }acc & $\sigma^{2}$ & avg\underline{ }acc & $\sigma^{2}$ & avg\underline{ }acc & $\sigma^{2}$ & avg\underline{ }acc & $\sigma^{2}$ & avg\underline{ }acc & $\sigma^{2}$  & Settings & avg\underline{ }acc & $\sigma^{2}$ & Dimensions \\ 
 \hline
                USPS & 0.7274&0.0049 & 0.7880&0.0040&  0.7271&0.0049& 0.8247&0.0017 &0.8358 & 0.0027 &0.8100 & 0.0019&0.8729&0.0012& 0.7984 & 0.0020 & 30
                &\textbf{0.8813} & 0.0018 & 30\\
                GLIOMA & 0.5667 & 0.0422 & 0.5330&0.0089 & 0.6000&0.0267&0.6667&0.0022&0.8333&0.0022&0.8000&0.0000&0.7000&0.0200& 0.5667 & 0.0356 & 50 & \textbf{0.8467} & 0.0167 & 41 \\
                lung & 0.7724&0.0069  & \textbf{0.9268} &0.0000 & 0.7724&0.0069  & 0.8374&0.0057  & 0.8293&0.0028  & 0.8374&0.0089  & 0.8618&0.0085  & 0.7724 & 0.0085 &50    & 0.8972 & 0.0000 &  50 \\
                madelon & 0.7237&0.0098 & 0.7365&0.0083 & 0.7218&0.0098 & 0.5949&0.0001 & 0.6538&0.0020 & 0.6724&0.0030 & 0.6359&0.0017 & 0.6699 &0.0045 & 50 & \textbf{0.7788} & 0.0099 & 15 \\
              COIL20 & 0.8576&0.0063 & 0.7535&0.0085 & 0.8322&0.0122 & 0.9051&0.0032 & 0.8935&0.0051 &0.8958&0.0041 &0.9363&0.0019 & 0.9051 & 0.0022 & 50
 & \textbf{0.9488} & 0.0022 & 50\\
                            TOX\_171     & 0.5333&0.0089 & 0.6095&0.0024 & 0.5143&0.0136 & 0.6857&0.0087 & 0.6190&0.0094 & 0.7048&0.0067 & 0.7143&0.0087 & 0.6571 & 0.0169 & 50 & \textbf{0.7214} & 0.0046 & 50 \\
                            
               warpPIE10P   & \textbf{0.8810} &0.0045 & 0.8333&0.0015 & \textbf{0.8810} &0.0045 & 0.8413&0.0066 & 0.7460&0.0054 & 0.7460&0.0168 & 0.8571&0.0034 & 0.8333 & 0.0049 & 50  & 0.8679 & 0.0013 & 50 \\
               
                Prostate\_GE & 0.9048&0.0045 & 0.9048&0.0045 & 0.9048&0.0045 & 0.9206&0.0020 & 0.8889&0.0005 & 0.8889&0.0081 & 0.9048&0.0045 & 0.9365&0.0005  & 50 & \textbf{0.9512}& 0.0005 & 35 \\
\hline
mean & 0.7459 & 0.0110 & 0.7607 & 0.0048 & 0.7442 & 0.0104 & 0.7846 & 0.0038 & 0.7875 & 0.0038 & 0.7944 & 0.0062 & 0.8104 & 0.0062 & 0.7674 &0.0094  & 47.5000 &\textbf{0.8617}& 0.0046 & 40.1250\\
\hline
\end{tabular}
}
\end{table*}

%基于子集方法的对比:SOTA+std+dimen
\begin{table*}[]
\centering
\caption{\textbf{State-of-the-art comparison for feature subset evaluation strategy.} Methods employing feature subset evaluation strategies focus on the utility of the feature sets. The 'dimensions' represents the number of features selected by methods. The meaning of 'mean', 'avg\underline{ }acc', and '$\sigma^{2}$' is the same as in Table.~\ref{SOTA-ranking}. The best accuracy in each row is marked with bold. Our method(MVMR-FS) achieves the best performance on all datasets.}
\label{SOTA-subset}
\resizebox{\textwidth}{!}{
    \begin{tabular}{c|ccc|ccc|ccc}
\hline
\multirow{2}*{Dataset} &\multicolumn{3}{c|}{CFS} & \multicolumn{3}{c|}{FCBF} &\multicolumn{3}{c}{ours model(MVMR-FS)}\\ 
 & avg\underline{ }acc & $\sigma^{2}$ & dimensions & avg\underline{ }acc & $\sigma^{2}$ & dimensions &avg\underline{ }acc &$\sigma^{2}$ &dimensions\\ \hline
                USPS & 0.8792 &0.0016 & 30 &0.8532 & 0.0036 & 26 & \textbf{0.8813} &0.0018 & 30 \\
                GLIOMA & 0.7000 & 0.0467 & 39 & 0.8333 & 0.0022 & 43 & \textbf{0.8467} & 0.0167 & 41 \\
                lung & 0.8618 & 0.0037 & 50  & 0.7967 & 0.0065 & 50 & \textbf{0.8972} & 0.0000 &50  \\
                madelon  & 0.6699&0.0022 &11 & 0.5564&0.0007 & 5 & \textbf{0.7788} &0.0099 & 15\\
              COIL20 & 0.9178&0.0042 & 50 & 0.9271 & 0.0032 & 50 & \textbf{0.9488} &0.0022 &50\\
                            TOX\_171 & 0.6571&0.0169 & 50 & 0.6667&0.0181 &50 & \textbf{0.7214} &0.0046  & 50 \\
                            
               warpPIE10P  & 0.7937 & 0.0160 & 50 & 0.8016 & 0.0190 & 32& \textbf{0.8679} & 0.0013 &50 \\
               
                Prostate\_GE  & 0.9206&0.0005 & 19 & 0.9365&0.0020 & 11 & \textbf{0.9512}&0.0005 &35\\ 
\hline
mean & 0.8000 & 0.0115 & 37.3750 & 0.7964 & 0.0069 & 33.3750 & \textbf{0.8617} & 0.0046 &40.1250 \\
\hline
\end{tabular}
}
\end{table*}

To illustrate the performance of our method, we compare MVMR-FS with the ten methods shown in Table.~\ref{baselines}. Table.~\ref{SOTA-ranking} shows the results of comparing MVMR-FS with individual feature evaluation methods. Table.~\ref{SOTA-subset} shows the results of comparing MVMR-FS with feature subset evaluation methods.\par
Compared with individual feature evaluation methods, MVMR-FS achieved the best performance in six of the eight datasets while also achieving the highest average accuracy. Our model not only achieved the highest accuracy on the GLIOMA, madelon, and Prostate\_GE datasets, but also further reduced the dimensionality of the feature subsets.  Although MVMR-FS does not select a lower-dimensional feature subset in each datasets, MVMR-FS does not require the user to specify the number of features.\par
MVMR-FS did not achieve the best performance on the lung and warpPIE10P datasets. ReliefF achieved the best performance on the lung dataset. Fisher Score and Trace Ratio achieved the best performance on the warpPIE10P dataset. We find that the best performance on all experimental datasets is achieved by methods that can be applied to continuous data without discretization. Therefore, using the method without discretization on continuous data can select a better feature subset.\par
Compared with the subset evaluation methods, MVMR-FS achieves the best performance in all datasets. The average dimension of FCBF is the smallest, but FCBF does not consistently achieve high accuracy on low-dimensional datasets. FCBF and CFS selected 5 and 11 features in madelon data, respectively. However, FCBF and CFS achieved particularly low accuracy in the madelon dataset, which were 0.5564 and 0.6699, respectively. Feature selection for FCBF and CFS in the madelon dataset is a failure.\par
Compared with the ten methods in Table.~\ref{SOTA-ranking} and Table.~\ref{SOTA-subset}, MVMR-FS achieves the highest average accuracy and improves the accuracy by 5\% to 11\%. The Fisher Score, ReliefF, and Trace Ratio are ranked in the last three in terms of average accuracy. These three methods can only measure the relevance of features, but cannot measure the redundancy. As a result, the performance is lower than methods that can measure both relevance and redundancy. MVMR-FS, CMIM, and CFS are ranked in the top three, and both MVMR-FS and CFS use feature subset evaluation strategies. Therefore, it is easier to find excellent subsets by measuring the feature set as a whole.\par

\subsection{Parameter analysis}
\label{canshufenxi}
The purpose of this experiment is twofold: 1. To analyze whether the bandwidth causes MVMR-FS to perform poorly in the lung and warpPIE10P. 2. To analyze the effect of bandwidth on the performance of MVMR-FS.\par 
\begin{figure*}[ht] 
    \centering    
    \subfigure[Parameter analysis on the lung] 
    {
        \label{lung}     
        \includegraphics[width=0.47\textwidth]{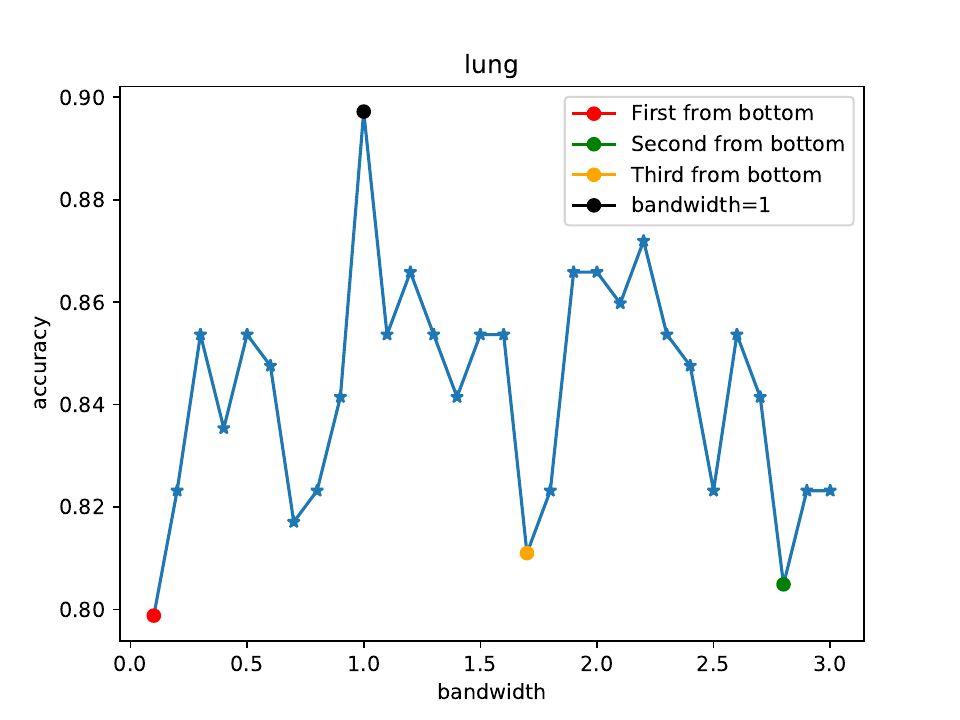}  
    }     
    \subfigure[Parameter analysis on the warpPIE10P] 
    { 
        \label{warpPIE10P}     
        \includegraphics[width=0.47\textwidth]{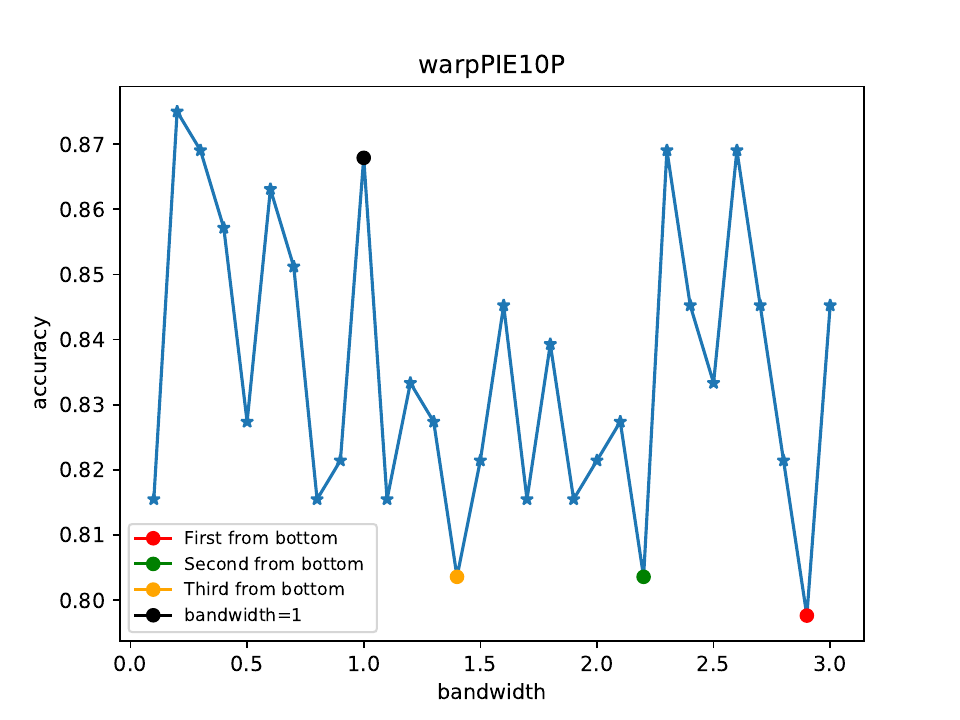}    
    }
    
    \subfigure[Parameter analysis on the TOX\_171] 
    { 
        \label{TOX}     
        \includegraphics[width=0.47\textwidth]{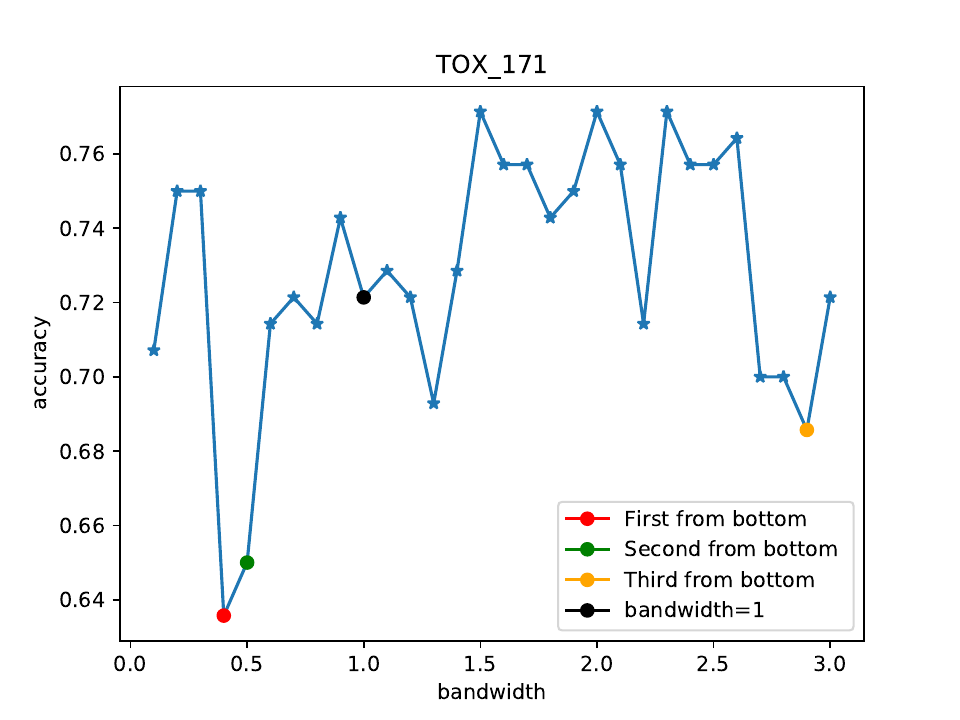}    
    }
    \subfigure[Parameter analysis on the madelon] 
    { 
        \label{madelon}     
        \includegraphics[width=0.47\textwidth]{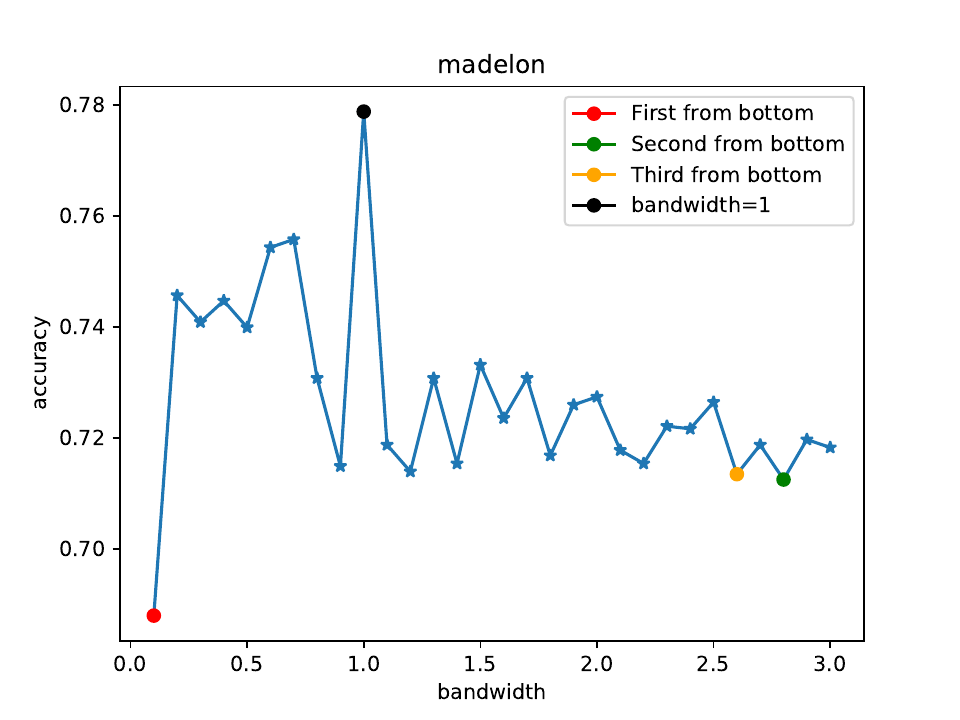}     
    }
\caption{\textbf{Results of the parametric analysis.} The black dots in the Figs.~\ref{lung},~\ref{warpPIE10P},~\ref{TOX},~\ref{madelon} indicate the accuracies corresponding to the bandwidth parameter used in this paper. We find that MVMR-FS usually achieves acceptable performance when the bandwidth is taken in the open interval (0.5,2.5).}     
\label{param-analysis-fig}     
\end{figure*}
We sampled the bandwidth at 0.1 intervals within the closed interval [0.1,3]. We then analyze the performance of MVMR-FS when the bandwidth is set to the values corresponding to the sampling points. The experimental results are shown in Fig.~\ref{param-analysis-fig}.\par
We found that the best accuracy of MVMR-FS on the lung dataset is the 0.8972 already achieved in Experiment.~\ref{results-SOTA}. The best accuracy of MVMR-FS for the warpPIE10P data is 0.8750, which is still lower than the accuracy achieved by Fisher Score. Thus, the main reason for the poor performance of MVMR-FS on the lung and madelon datasets is not the choice of bandwidth.\par 
In this experiment, MVMR-FS performs poorly when the bandwidth is taken in the interval [0.1,0.5] or [2.5,3.0]. The bandwidth parameters corresponding to the lowest accuracies on all four datasets lie in the above interval. Furthermore, most bandwidth values corresponding to the penultimate and antepenultimate accuracies are also within the above intervals. \par
The reason is that when the bandwidth falls within the above interval, the probability density curve obtained through kernel density estimation becomes either overly smooth or steep, failing to reflect the original distribution of the data. This leads to MVMR-FS being unable to accurately assess the relevance and redundancy among features, resulting in the selection of non-optimal features.\par
We find that MVMR-FS usually achieves acceptable performance when the bandwidth is taken in the open interval (0.5,2.5). But we cannot set the bandwidth parameter too large or too small.\par
\section{Conclusion and future work}
\label{Conclusion}
In this paper, we propose MVMR-FS for solving the problem that filter-based FS cannot measure redundancy in continuous data and requires human involvement. MVMR-FS comprises three key components: SKDE, UKDE, and the MVMR criterion. SKDE and UKDE offer perspectives on the similarities and dissimilarities among various features based on probability density, enabling MVMR-FS to directly quantify redundancy for continuous data. MVMR utilizes the information provided by SKDE and UKDE to assess the global optimality of feature subsets, thereby eliminating the dependence on manually specifying the number of features. The experimental results demonstrate that MVMR-FS significantly outperforms state-of-the-art methods and achieves the best results on benchmark datasets. \par
However, MVMR-FS also has some limitations. The bandwidth parameter in kernel density estimation can affect the performance of MVMR-FS. In the TOX\_171 dataset of Experiment~\ref{canshufenxi} , the difference between the best and the worst performance is more than 12\%. Hence, how adaptively choosing the optimal bandwidth parameter in MVMR-FS is a worthwhile problem. This is also one of the directions of our future work.

\section*{Acknowledgements}
This research was supported by the National Key Research and Development Program of China(2020YFC2008501).

%%%%%%正文%%%%%%%%
\printcredits

\bibliographystyle{model1-num-names}

% Loading bibliography database 参考文献
\bibliography{reference}

\end{document}